\documentclass[10pt,twocolumn,letterpaper]{article}

\usepackage[pagenumbers]{cvpr} 

\usepackage[dvipsnames]{xcolor}
\usepackage{tikz}
\usepackage{float} 
\usepackage{url}

\definecolor{cvprblue}{rgb}{0.21,0.49,0.74}
\usepackage[pagebackref,breaklinks,colorlinks,citecolor=cvprblue]{hyperref}

\def\paperID{4444} 
\def\confName{CVPR}
\def\confYear{2024}

\title{PhytNet - Tailored Convolutional Neural Networks for Custom Botanical Data}

\author{Jamie R. Sykes\\
Department of Computer Science, University of York\\
Deramore Lane, York, Yorkshire, UK. YO10 5GH\\
{\tt\small jamie.sykes@york.ac.uk}
\and
Katherine J. Denby\\
Centre for Novel Agricultural Products, Department of Biology, University of York\\
Wentworth Way, York\\
{\tt\small katherine.denby@york.ac.uk}
\and
Daniel W. Franks\\
Department of Biology, University of York\\
Wentworth Way\\
{\tt\small daniel.franks@york.ac.uk}
}

\begin{document}
\maketitle
\begin{abstract}
Automated disease, weed and crop classification with computer vision will be invaluable in the future of agriculture. However, existing model architectures like ResNet, EfficientNet and ConvNeXt often underperform on smaller, specialised datasets typical of such projects. We address this gap with informed data collection and the development of a new CNN architecture, PhytNet. Utilising a novel dataset of infrared cocoa tree images, we demonstrate PhytNet's development and compare its performance with existing architectures. Data collection was informed by analysis of spectroscopy data, which provided useful insights into the spectral characteristics of cocoa trees. Such information could inform future data collection and model development. Cocoa was chosen as a focal species due to the diverse pathology of its diseases, which pose significant challenges for detection. ResNet18 showed some signs of overfitting, while EfficientNet variants showed distinct signs of overfitting. By contrast, PhytNet displayed excellent attention to relevant features, no overfitting, and an exceptionally low computation cost (1.19 GFLOPS). As such PhytNet is a promising candidate for rapid disease or plant classification, or precise localisation of disease symptoms for autonomous systems.
\end{abstract}    
\section{Introduction}
\label{sec:intro}
\subsection{Background}
Computer vision projects in areas like plant pathology and agronomy often have limited data for training and are thus constrained in choice of model architecture. ResNet \cite{heDeepResidualLearning2016a}, EfficientNet \cite{tanEfficientNetRethinkingModel2020a} and ConvNeXt \cite{liuConvNet2020s2022} variants are among the best current neural networks available for such smaller datasets \cite{sykesComputerVisionPlant2023}. However, these models were developed for general image recognition tasks and are honed to perform well on huge benchmark datasets such as ImageNet, which has 1.4M images and 1,000 classes \cite{wooConvNeXtV2CoDesigning2023}. The focus in recent years of model development has moved away from convolutional neural networks (CNNs), towards the larger transformer and CNN-transformer hybrid models such as DETR \cite{zhuDeformableDETRDeformable2021}. While this trend has yielded a lot of highly performant models, it leads to problems of scale for many real-world applications. Transformers such as ViT and larger CNNs like ConvNeXt scale exceptionally well with large datasets. However, training these models and curating the requisite massive training datasets is prohibitively expensive. Additionally, the increased runtime cost of these models necessitates fast internet connections and/or expensive hardware \cite{zhangMiniViTCompressingVision2022}, further prohibiting their use. The authors of ViT acknowledge such issues for models like Swin transformer \cite{liuSwinTransformerHierarchical2021a} and V-MoE \cite{riquelmeScalingVisionSparse2021}, which have 3 and 14.7 billion parameters respectively. However, their solutions, mini-ViT \cite{zhangMiniViTCompressingVision2022} and tiny-ViT \cite{wuTinyViTFastPretraining2022a} still have tens of millions of parameters. 

\subsection{Fitting architectures to datasets}
In projects with small training datasets, it's crucial to match the neural network's size to the size of the dataset. This will help prevent overfitting and promote effective generalisation. If this is not done, the model may memorises the training data, including its noise and outliers, rather than learning genuine patterns. This will result in poor performance on new data and give a misleading sense of model accuracy based on performance metrics observed during training. 

In response to this, we took inspiration from the design of ResNet, ConvNeXt and EfficientNet to develop a state-of-the-art model, named PhytNet, and an accompanying hyperparameter search. The aim was to produce a CNN that would perform well on realistically sized datasets and have minimal computation cost at runtime. CNNs have several intrinsic benefits over transformers that, for image classification, allow CNNs to outperform transformers while requiring fewer parameters and far less data for training \cite{liuConvNet2020s2022}. 

Many of the recent performance gains in image classification have come from novel training procedures involving semi-supervised and unsupervised pre-training \cite{sykesComputerVisionPlant2023, wooConvNeXtV2CoDesigning2023}. While we intend to apply such techniques to the pre-training of PhytNet with a custom plant dataset, that work is beyond the scope of this study and will be presented in a separate future study. Here we aim first to define a small, high quality dataset and use it to guide the development of a new model architecture. With PhytNet we bring both the best established features and latest advancements to small model design. In doing so we also report observations regarding the localisation of machine visible disease symptoms in cocoa trees and identify the most informative parts of the UV to IR electromagnetic spectrum for cocoa disease detection. While not central to the theme of this work, such results will be informative for development of future training datasets and models as we continue to search of informative signals of disease. Lastly we apply cross-validation and gradient based class activation maps (Grad-CAM) \cite{selvarajuGradCAMVisualExplanations2017} to compare and contrast PhytNet against ResNet18, EfficientNet-b0, EfficientNet-V2s and ConvNeXT tiny.

\subsection{PhyNet development}

To develop PhytNet, and demonstrate the practical means of its development, we use the real-world challenge of classifying images of disease in cocoa trees, \emph{Theobroma cacao}. Cocoa is highly vulnerable to fungal and oomycete diseases due to the humid conditions in which it grows. Most disease control in cocoa involves costly manual search and phytosanitation \cite{meinhardtMoniliophthoraPerniciosaCausal2008}. Since most cocoa farmers live below the poverty line, they can't easily afford such expenses or chemical pesticides \cite{boeckxPovertyClimateChange2020}. Cocoa therefore is a prime use case for automated disease detection and mapping. Additionally, cocoa's diversity of diseases poses a significant challenge for human and machine disease classifiers. This makes it an ideal challenge by which to guide the development of a phytology focused classification model. Each year, about 38\% of the global cocoa crop is lost to three main diseases: black pod rot (BPR; \emph{Phytophthora palmivora}), witches' broom disease (WBD; \emph{Moniliophthora perniciosa}) and frosty pod rot (FPR; \emph{Moniliophthora roreri}) \cite{gidoinShadeTreeSpatial2014a}. These three diseases, which impose a significant burden on cocoa farmers and, indirectly, on the natural environment \cite{malhiClimateChangeDeforestation2008, kuokhoSystematicReviewSlashandBurn2020}, will be of central concern in this study. 

\section{Methods}
\label{sec:formatting}

\subsection{Spectroscopy}

We used a MultispeQ v2.0 \cite{kuhlgertMultispeQBetaTool2022} to measure photosystem II quantum yield (Phi2) and non-photochemical quenching (NPQt) in cocoa trees with different disease states. Both Phi2 and NPQt have been shown to have significant negative and positive correlations with disease index respectively \cite{kuhlgertMultispeQBetaTool2022}. These measurements were taken to assess if non-visible signals of disease could be detected in the foliage of cocoa trees. Readings were taken from 1) healthy leaves on trees with no signs of disease nearby (n=9), 2) leaves clearly infected with WBD (n=5) and 3) the nearest leaf to pods or leaf tissue infected with BPR (n=10), FPR (n=5) or WBD (n=5). Measurements were taken across four sites northeast to southeast of Guayaquil, Ecuador. 

To measure the reflectance spectrum of diseased and healthy cocoa pods, we used a field spectrometer, produced by tec5 (Steinbach, Germany). This spectrometer had a receptive range of 310 nm-1100 nm. The spectroscopy data was gathered on four farms located east of Guayaquil, Ecuador, near the foothills of the Andes mountains. Readings were taken in accordance with the manufacturers instructions. We trained a random forest classifier to predict disease state from the reflectance data to estimate the relative importance of different spectral bands. Hyperparameter values for the random forest were chosen using a grid search with 5 fold cross-validation. The test accuracy of this classifier was 78\%. We considered only healthy cocoa pods (n=47) and pods with clear symptoms of BPR (n=40) or FPR (n=24). No WBD was present in the cocoa pods at these plantations due to the time of year.

\subsection{Image data collection}

We collected two datasets of infrared (n=240) and RGB (n=240) images taken concurrently with the same Olympus OMD EM-5 II camera. While the IR data was collected primarily for the purposes of model development, the additional RBG data allowed us to compare an IR and RGB trained ResNet18 model for disease classification. Images were of healthy cocoa trees bearing pods as well as trees bearing pods with BPR, FPR and WBD in equal numbers. Diseased pods showed clearly visible and easily diagnosable symptoms in the early to mid stages of disease development.
We randomised across a variety of factors such as geographic location, disease stage, crop variety and air temperature. The two datasets were collected with and without a 720 nm neutral density filter to block all visible light. For the RGB images, the factory camera settings were used with the "intelligent-auto" program selected. A tripod was used to collect the IR images to allow for longer exposure and the following camera settings were applied. ISO: 200, white balance: 2000 kelvin, noise reduction: On, noise filter: low, shutter speed: auto, delay: 12 seconds.   
These images were collected at two research stations, either side of the Andes. The research stations, belonging to Instituto Nacional de Investigaciones Agropecuarias (INIAP), were located at Pichilingue and Coca, Ecuador. 
\subsection{Model development and optimisation}
PhytNet is designed to be easily configured for specific datasets and its design choices reflect a balance between overfitting, performance and computational efficiency. It is composed of a series of bottle neck blocks, using group normalisation in place of batch normalisation. Unlike batch normalisation, which normalises across the batch dimension, group normalisation divides the channels into groups and normalises within each group. This can be particularly useful for smaller batch sizes and is a deviation from the standard ResNet design. Batch normalisation has also been shown to be sub-optimal for plant pathology in particular \cite{gongAnalyzeCOVID19CT2021a,sykesComputerVisionPlant2023}.
In PhytNet various parameters, described below, are determined by a configuration file. This allows the model configuration to be easily optimised for a given dataset with a guided optimisation sweep. Similar to EfficentNet, the network concludes with a global average pooling layer before a fully connected layer, which allows for easy scaling and variable image input size post training. 

Starting with a basic residual CNN and the intentionally small dataset of 240 IR images, a series of training runs were executed while various configurations were manually tuned and tracked using the Weights and Biases platform (wandb; San Francisco, California, USA) \cite{biewaldExperimentTrackingWeights2020}. These configurations included different activation functions such as RELU and GELU \cite{hendrycksGaussianErrorLinear2023}, attention mechanisms including multi-head attention \cite{vaswaniAttentionAllYou2017} and squeeze-excitation layers \cite{huSqueezeandExcitationNetworks2019}, dimension reduction such as max pooling, average pooling and global average pooling, convolution block/bottleneck block configurations, stochastic depth \cite{huangDeepNetworksStochastic2016}, dropout \cite{hintonImprovingNeuralNetworks2012}, batch normalisation \cite{ioffeBatchNormalizationAccelerating2015b}, layer normalisation \cite{baLayerNormalization2016}, group normalisation \cite{wuGroupNormalization2018a}, model depths and dense layer configurations. Any adjustments to the architecture were retained if they improved the validation F1 score while reducing, or at least maintaining, signs of overfitting. In addition to performance metrics such as accuracy, F1 and loss, the number of trainable parameters and GFLOPS were also recorded for each model. Choosing a model with a high valuation F1 score, comparatively low training F1 and the minimum number of trainable parameters and GLOPS would help to avoid overfitting and reduce computational cost. This strategy of model choice was also applied in the subsequent optimisation sweep.

Once the architecture was chosen, an optimisation sweep was run using wandb's Bayesian optimisation method to optimise for validation set F1. Optimising for validation F1 in this way should act to reduce overfitting as we searched for a model that performed best on the validation set, not the training set. The Bayesian optimisation strategy employed here is a sequential process that is particularly useful for optimisation of the hyperparameters of models that are expensive to train, such as neural networks. In this case, Bayesian optimisation uses a Gaussian process to model the validation F1 as a function of model hyperparameters. The Gaussian process gives a distribution over functions that describes the relationship between hyperparameters and F1. The process starts with a prior based on a mean function and a covariance function. Then, as the process observes new data points from previous trained models, it updates its beliefs about the validation F1 function and generates new hyperparameters based on metrics such as expected improvement, probability of improvement and upper confidence bound. Though in the wandb documentation, it is not specified which is used in this case.

This sweep optimised the kernel size of the middle convolutional layer of each bottleneck block (1:19), number of convolution layer channels (16:128), number of bottleneck blocks in each convolution block (1:4), square image input size (200:500 pixels), learning rate (1\textsuperscript{-6}:1\textsuperscript{-3}), number of output channels (4:10), and beta1 (0.88:0.99) and beta2 (0.93:0.999) values of the AdamW optimiser, which control the exponential moving average of weight updating. These values were selected to facilitate the search for a model that would avoid over fitting, train in a controlled manner while allowing enough stochasticity to properly search the loss landscape, and have an appropriate kernel size to focus its attention on the features of the given dataset. Importance of kernel size optimisation is exemplified in ConvNeXt \cite{liuConvNet2020s2022}.

During the sweep, models larger than 6 GFLOPS or two million trainable parameters were terminated before training. 645 models were trained using one NVIDIA GeForce GTX 1080 Ti GPU and one NVIDIA Quadro P6000 GPU. Each model was trained on a single GPU, taking approximately 4 minutes to train. Early stopping was applied to halt training when the validation loss failed to decrease for 20 consecutive epochs, at which point the checkpoint with the best validation F1 was saved. Simple supervised training was used with AdamW optimisation (weight decay=1\textsuperscript{-4} and eps=1\textsuperscript{-6}). L1 regularisation was added to the cross entropy loss function by summing the value of all model parameters and multiplying that value by a weight of 1\textsuperscript{-5}. The following image augmentations were randomly applied during training: horizontal flip, Gaussian blur and random rotation of 0-5 degrees. We also optimised for the number of output nodes and observed that the best performing models had 7 or 8 output nodes, despite having only 4 image classes. Potential explanations for this observation are given in the discussion section.

\subsection{Model evaluation}
To asses the performance of PhytNet, we compared it to four competing architectures; ResNet18, EfficientNet-b0, EfficentNet-V2s and ConvNeXt tiny. Each of these four architectures was trained using the procedure described above with the addition of an optimisation sweep for image input size, learning rate, and beta1 and beta2 values of AdamW. This allowed for a fair comparison between models, giving the competing architectures a chance to avoid overfitting and produce favourable results. To ensure reputability, torch backends were set to deterministic, torch random seeds were set to 42 and the data loader random seeds were set based on worker ID. Initial weights were generated using PyTorch default methods rather than using pre-trained ImageNet weights to ensure a fair comparison between models as no ImageNet weights are available for PhytNet yet. Plant dataset specific pre-trained weights for PhytNet will be produced in a subsequent study.  

The performance metrics of the four models were estimated using 10-fold cross validation, with the same data split for each model. While 10\% of the data was reserved for validation during model development, the typical 10\% test data was not used for evaluation for three reasons. 1) In this study, the images are highly consistent in their features, meaning that any subset would closely resemble the overall distribution of the training data, leading to a risk of pseudo-replication. 2) This dataset is too small to justify discarding 10\% of the training data for testing. 3) While cross-validation is more computationally expensive, it has been shown to be a more robust means of model testing. This is because it gives a more conservative estimate of model performance, while better reflecting real world conditions. It also allows for the assessment of the distributions of performance metrics, which can be used to detect overfitting \cite{kingCrossvalidationSafeUse2021}. 

Finally, using Grad-CAM \cite{selvarajuGradCAMVisualExplanations2017}, class activation maps were produced to inspect the informative features used by each model. Grad-CAM allows for the identification of areas of an image that are used by a CNN to inform decision making. It identifies the class of interest and computes the gradients of the class score with respect to feature maps produced from the final convolutional layer output. This information is expressed as a heatmap, which is superimposed on to the original image. Using this method allowed us to further asses the degree of overfitting in each model and catch naive behaviour that is not apparent in summary statistics.  


\section{Results}
\subsection{Locating machine visible symptoms}
\begin{figure}
  \centering
  \begin{subfigure}{\linewidth}
    \includegraphics[width=\linewidth]{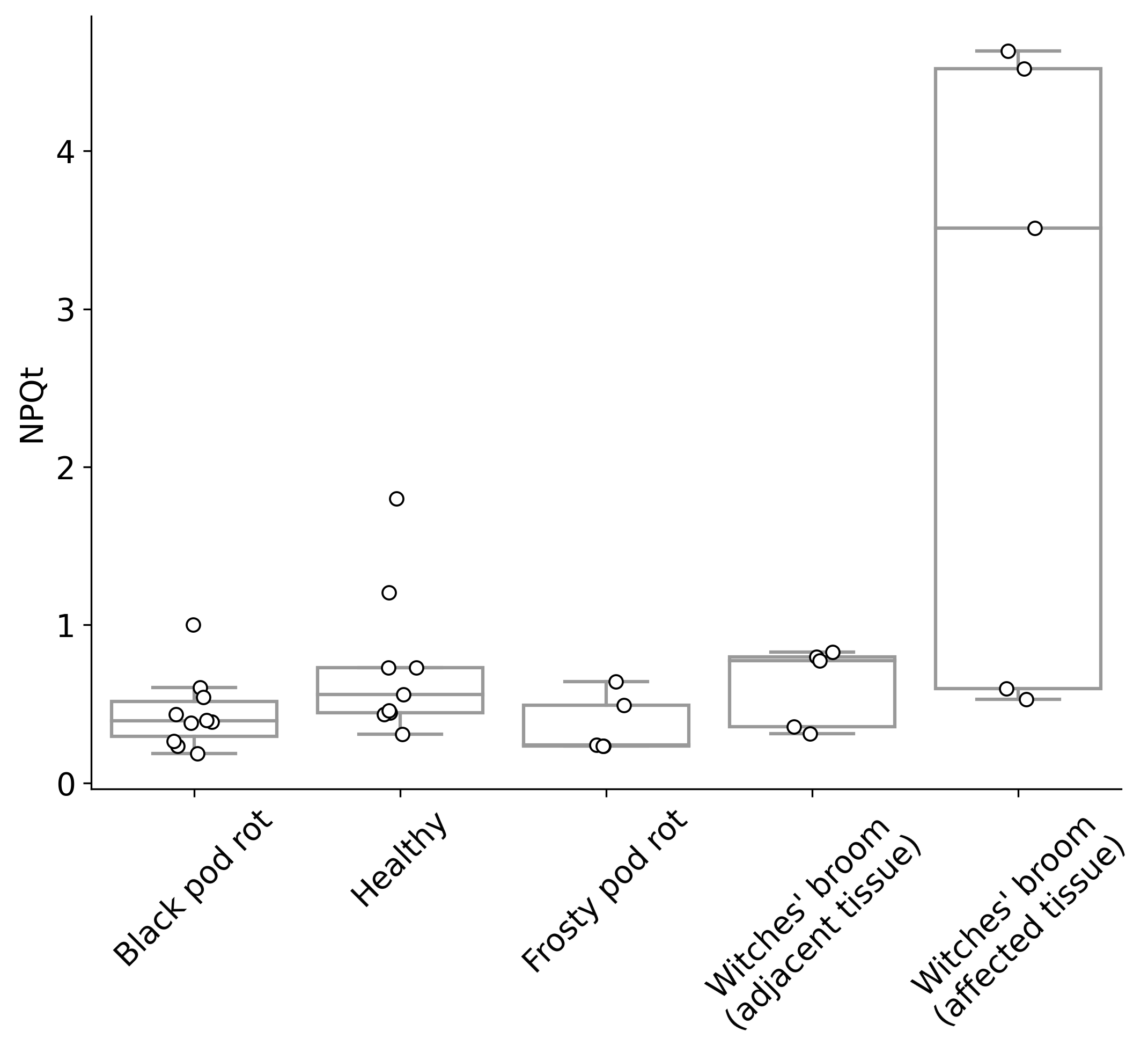}
    \caption{Non-photochemical quenching}
    \label{fig:MultiSpeQ-a}
  \end{subfigure}
  
  \vspace{1em} 
  
  \begin{subfigure}{\linewidth}
    \includegraphics[width=\linewidth]{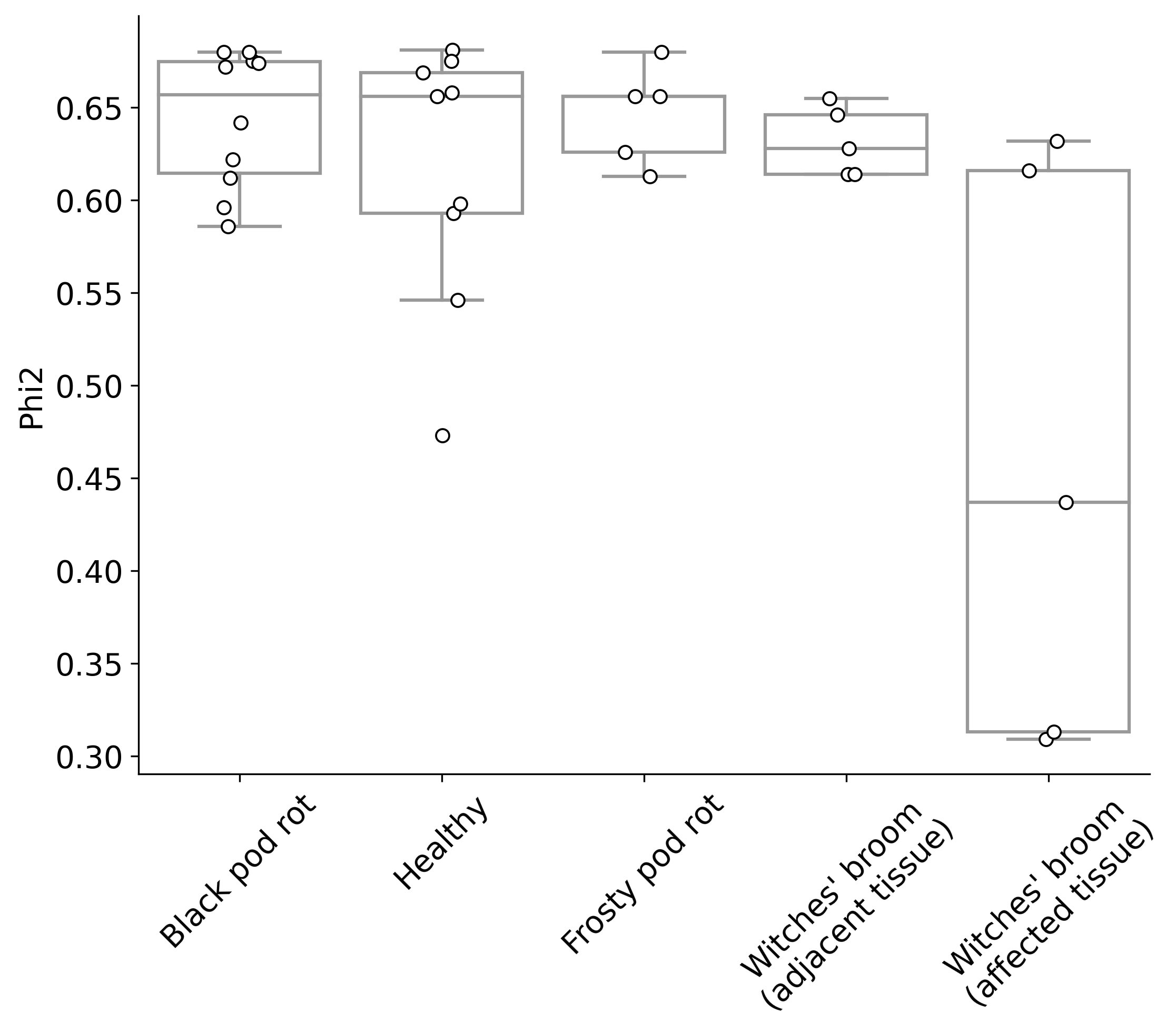}
    \caption{Quantum yield of Photosystem II}
    \label{fig:MultiSpeQ-b}
  \end{subfigure}
  
  \caption{Distributions of non-photochemical quenching (NPQt)(a) and photosynthetic yield (Phi2)(b) of cocoa trees with different disease states. Box plots show the interquartile range with whiskers at 1.5 times the IQR from the first and third quartiles. Raw data points plotted as white circles. Measurements taken from cocoa trees in five disease states with a MultispeQ v2.0. BPR n=10, FPR n=5, Healthy n=9, Witches' broom (adjacent tissue) n=5,  Witches' broom (affected tissue) n=5.}
  \label{fig:MultiSpeQ}
\end{figure}

\Cref{fig:MultiSpeQ} shows the result of measuring Phi2 and NPQt in diseased and healthy cocoa trees using a MultispeQ V2. These measurements were taken to asses if non-visible signals of disease could be detected in the foliage of cocoa trees.  While some of the leaves that were visibly infected with WBD showed clear, but inconsistent, signals of compromised photosynthesis, the leaves adjacent to pods and/or leaves infected with BPR, FPR or WBD showed no signs of reduced Phi2 or increased NPQt, relative to healthy trees. As such we see no evidence here that these diseases can be detected through compromised photosynthesis in foliage that shows no human visible signs of disease. 

\subsection{Identifying informative light spectra}
\begin{figure}
  \centering
  \begin{subfigure}{\linewidth}
    \includegraphics[width=\linewidth]{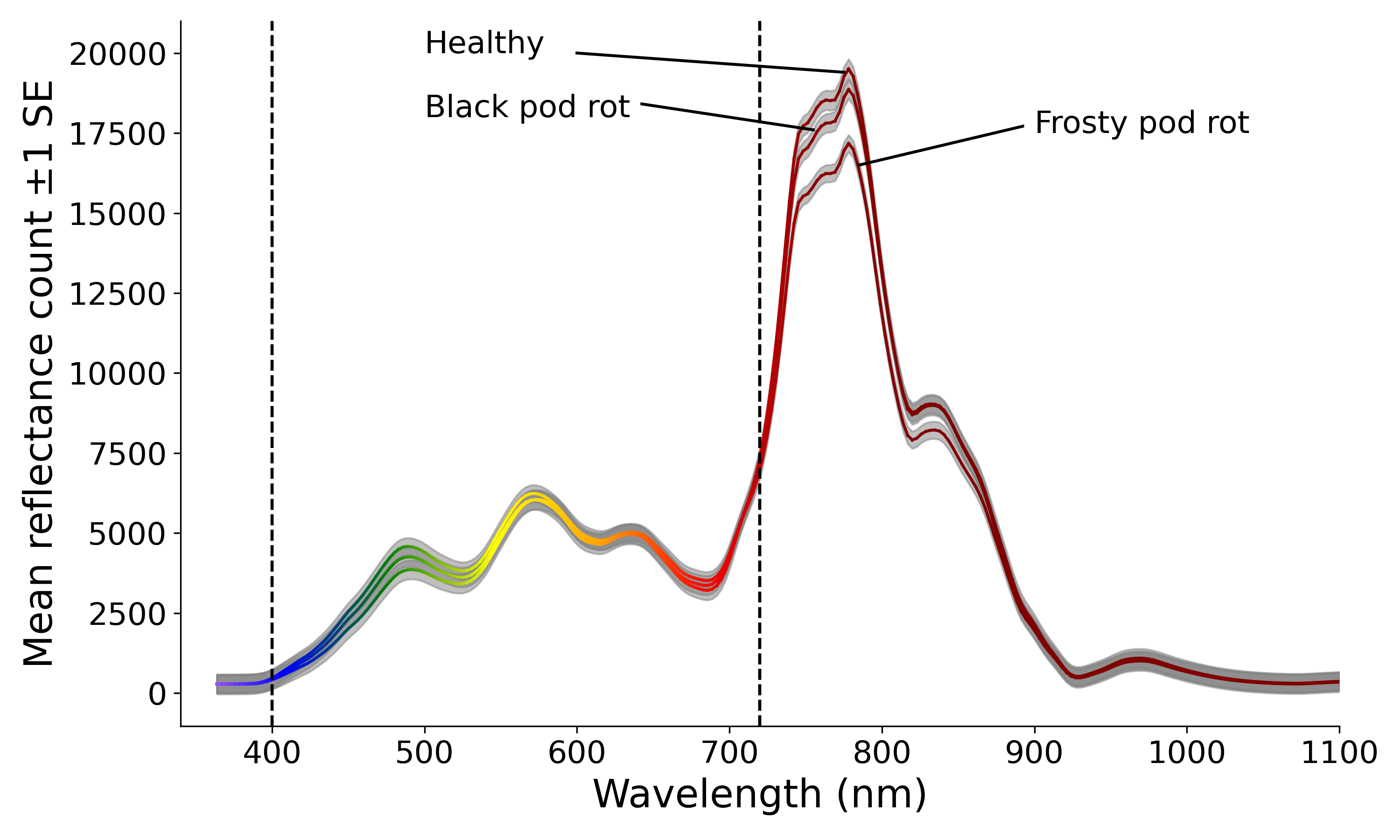}
    \caption{Mean reflectance spectrum measurements}
    \label{fig:spectroscopy-a}
  \end{subfigure}
  \vspace{1em} 
  \begin{subfigure}{\linewidth}
    \includegraphics[width=\linewidth]{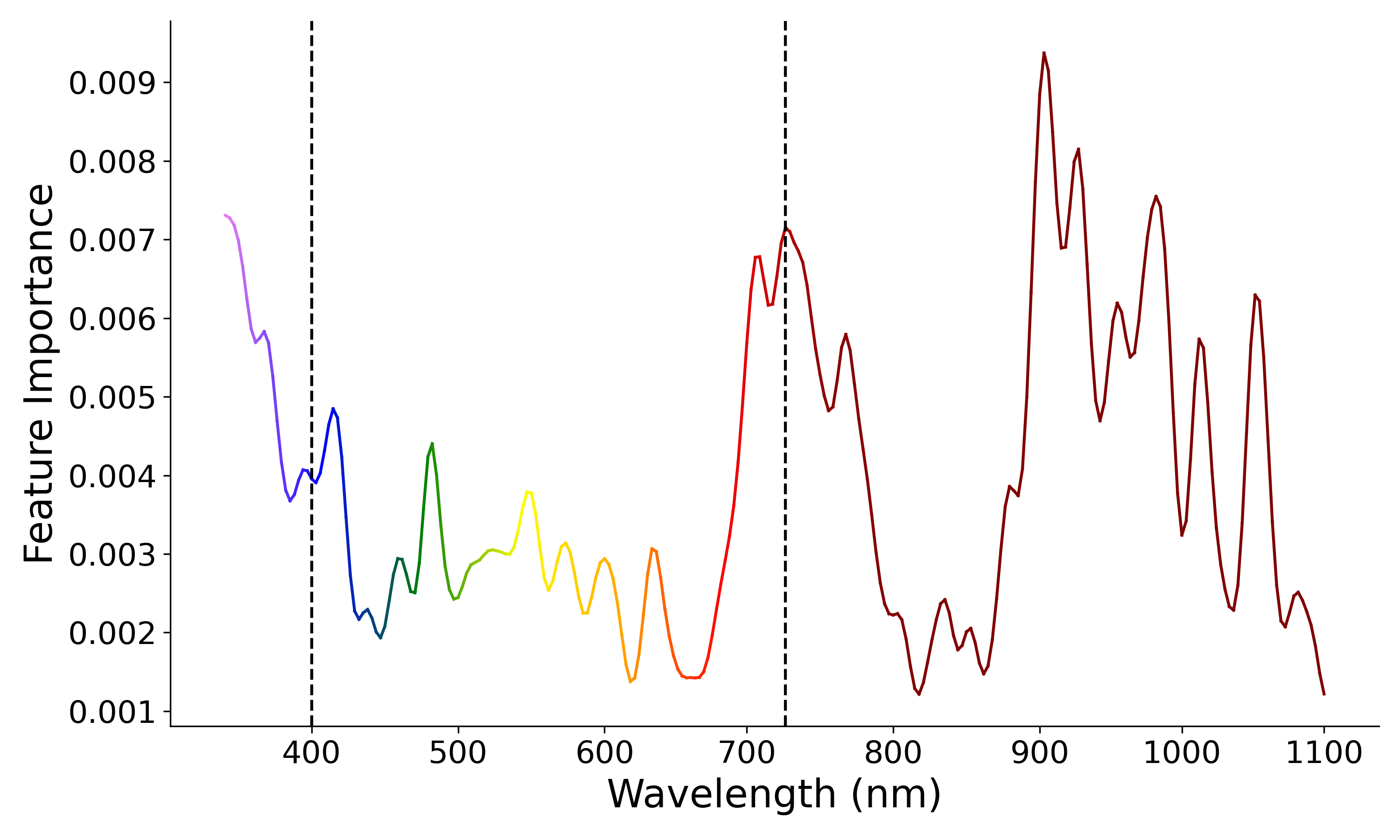}
    \caption{Feature importance scores}
    \label{fig:spectroscopy-b}
  \end{subfigure}
  \caption{Mean reflectance spectrum measurements, ±1 standard error (a) and feature importance scores (b) derived from a random forest classifier trained on the same data. Spectroscopy data were gathered from cocoa pods that were either healthy or showing early to mid-stage frosty pod rot or black pod rot symptoms. The standard error is shown by grey shading and the dotted lines at 400 and 720 nm show the bounds of the human visible spectrum. Random forest classifier train accuracy: 95.45\% and test accuracy: 78.26\%.}
  \label{fig:spectroscopy}
\end{figure}

\Cref{fig:spectroscopy} shows the mean reflectance spectrum of cocoa pods and the relative importance of each spectra in classifying disease. While the healthy and BPR infected pods seem to show no difference in their reflectance spectrum (\cref{fig:spectroscopy-a}), the FPR infected pods show a clear reduction in IR light reflectance. This may represent an informative signal to detect this disease, which can be asymptomatic until the late stages of development.

While IR light is among the most informative spectra (\cref{fig:spectroscopy-b}), the low importance scores shown here suggest that any single band of light spectra alone is not highly informative in classifying cocoa disease. However, the moderately high validation accuracy of this random forest classifier (78\%) confirms that, taken together, spectroscopy data are informative. \Cref{fig:spectroscopy-b} also shows that many of the most informative light spectra are in the UVA (below 400 nm) and infrared range (above 720 nm). \emph{i.e.} data that is mostly not captured in RGB images \cite{linharesHowGoodAre2020}. Additionally, the FPR per class accuracy (80\%) from the random forest shows that the dip in IR light of FPR is not sufficient alone to perfectly classify FPR, though the spike in feature importance between 700-800 nm suggests that it is informative. 

\subsection{Model evaluation}
\begin{table*}
\centering
\caption{Results of 10-fold cross validation analysis comparing PhytNet to four competing architectures, all trained on the same IR cocoa disease dataset. \emph{n.b.} GFLOPS were calculated using the optimised input size, shown here as pixels\textsuperscript{2}. GFLOP values reported by the original model authors may differ because of this.}

\begin{tabular}{|c|c|c|c|c|c|c|c|}
\hline
Model & Train F1 & Val F1 & Train loss & Val loss & GFLOPS & n pixels\textsuperscript{2} & n parameters \\
\hline
PhytNet & 0.60 & 0.61 & \bf{0.93} & \bf{1.17} & \bf{1.19} & 285 & \bf{336,196} \\
\hline
ResNet18 & \bf{0.62} & \bf{0.65} & 1.8 & 1.94 & 6.16 & 408 & 11,178,564 \\
\hline
EfficentNet b0 & 0.87 & 0.69 & 2.31 & 3.16 & 1.45 & 424 & 3,970,656 \\
\hline
EfficentNet V2s & 0.89 & 0.71 & 2.26 & 3.32 & 13.23 & 485 & 19,913,468 \\
\hline
ConvNeXt tiny & 0.41 & 0.52 & 1.82 & 1.88 & 3.77 & 212 & 27,813,508 \\
\hline
\end{tabular}
\label{tab:CrossVal_results}
\end{table*}

\begin{figure}
    \centering
    \begin{tikzpicture}
        \node[anchor=south west,inner sep=0] (image) at (0,0) {\includegraphics[width=\columnwidth]{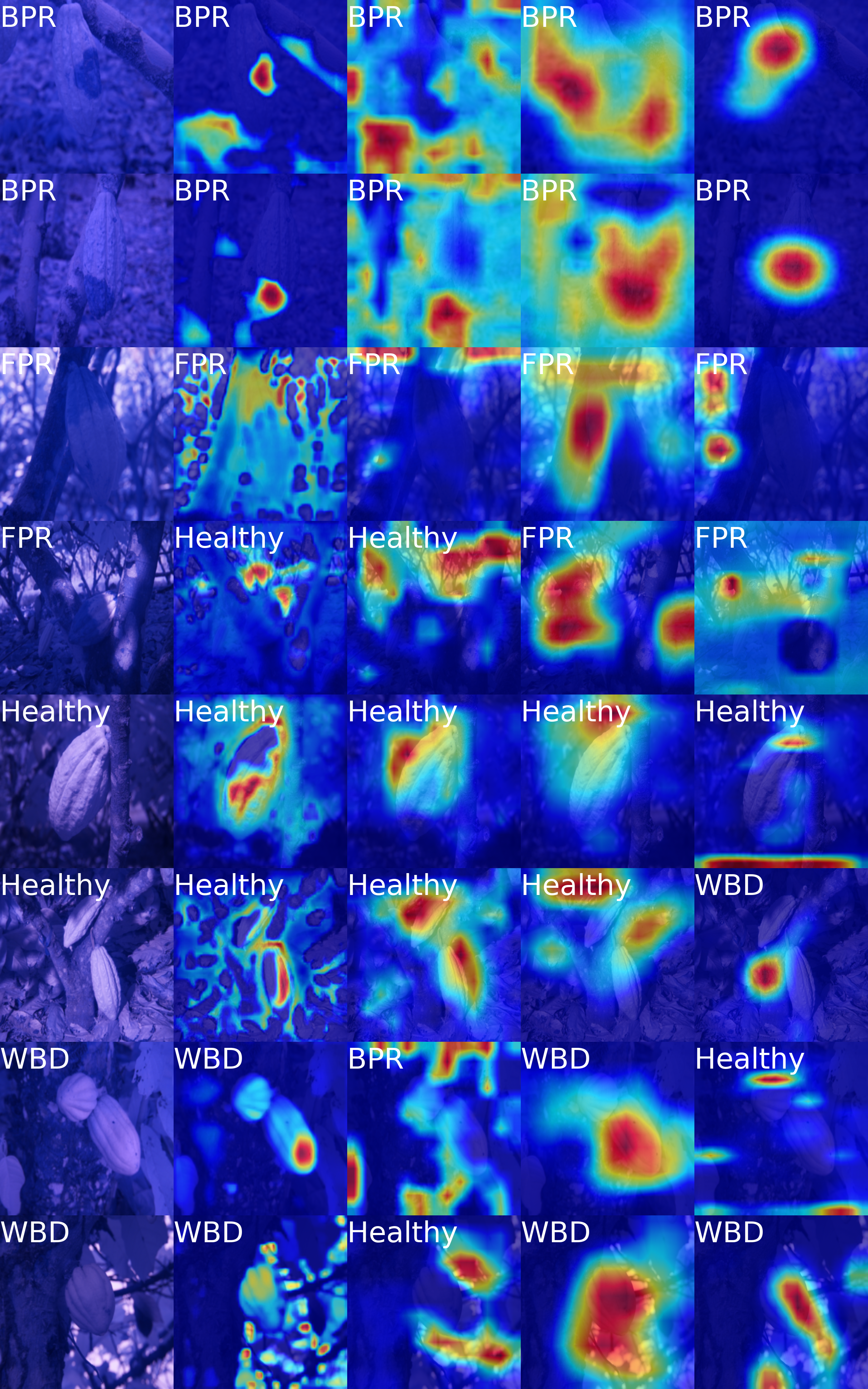}};
        \begin{scope}[x={(image.south east)},y={(image.north west)}]

            \node[anchor=south, font=\footnotesize] at (0.1,1) {Input};
            \node[anchor=south, font=\footnotesize] at (0.3,1) {PhytNet};
            \node[anchor=south, font=\footnotesize] at (0.5,1) {ResNet18};
            \node[anchor=south, font=\footnotesize] at (0.7,1) {Eff.Net-b0};
            \node[anchor=south, font=\footnotesize] at (0.9,1) {Eff.Net-V2s};

        \end{scope}
    \end{tikzpicture}
    \caption{Infrared images with class activation heatmaps produced using Grad-CAM and four CNNs. Models used are PhytNet, ResNet18, EfficeintNet-b0 and EfficientNetV2 (left to right). The leftmost column shows raw input images with ground truth labels in white, other white labels are predicted by each model.}
    \label{fig:Grad-CAM attention maps}
\end{figure}

\Cref{tab:CrossVal_results} shows that while EfficientNet-b0 and EfficientNet-V2s give a relatively high mean validation F1: 69\% (95\% CI: 0.64-0.74) and 71\% (95\% CI: 0.66-0.75) respectively, both have a mean training F1 18 percentage points higher than their mean validation F1, suggesting that they overfit to this data. This is corroborated by their higher validation loss than training loss and by the class activation maps produced using Grad-CAM (\cref{fig:Grad-CAM attention maps}). \Cref{fig:Grad-CAM attention maps} shows that EfficientNet-b0 focuses its attention poorly, except perhaps for the first WBD image. EfficeintNet-v2s seems to focus its attention quite well for the BPR images, thought in both cases it seems to focuses more on the healthy tissue than the disease lesions. Furthermore, EfficeintNet-v2s seems to fail completely to focus its attention on disease symptoms, or even the focal tree, in the other images, despite getting most classifications correct. This is a textbook definition of overfitting. \emph{i.e.} These models are utilising image features that are consistent in the training set but are unrelated to the actual disease symptoms, potentially leading to poor generalisation on new or unseen data. Additionally, at 13.23 GFLOPS, EfficeintNet-v2s is far more computationally expensive than the other models described here, making it inappropriate for rapid classification when used on edge devices. ConvNeXt tiny performed with poor F1 scores on this dataset (\Cref{tab:CrossVal_results}), though this is not surprising given its huge number of parameters (27.8 million) and the small size of this dataset. However, ConvNeXT tiny's loss values were similar to those of ResNet18 and, with an optimised input size of 212 pixels\textsuperscript{2}, its computational cost was measured at a  very low 3.77 GFLOPS. Though, owing to the low F1 scores, ConvNeXt is not considered in subsequent analyses here.

\begin{figure*}
  \centering
  \begin{subfigure}{0.65\linewidth}
    \includegraphics[width=\linewidth]{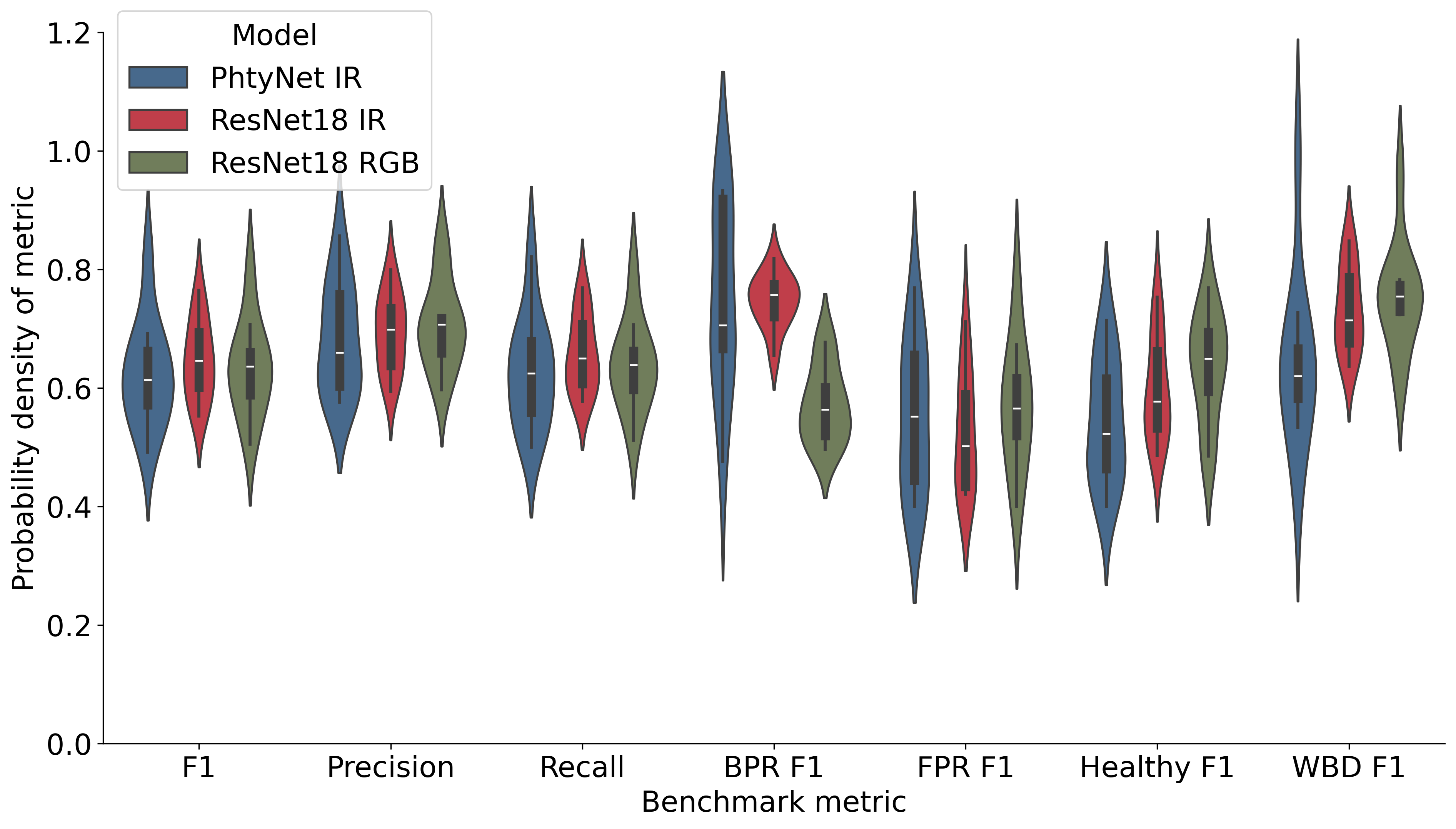}
    \caption{Training dataset results}
    \label{fig:short-a}
  \end{subfigure}
  \vspace{1em} 
  \begin{subfigure}{0.65\linewidth}
    \includegraphics[width=\linewidth]{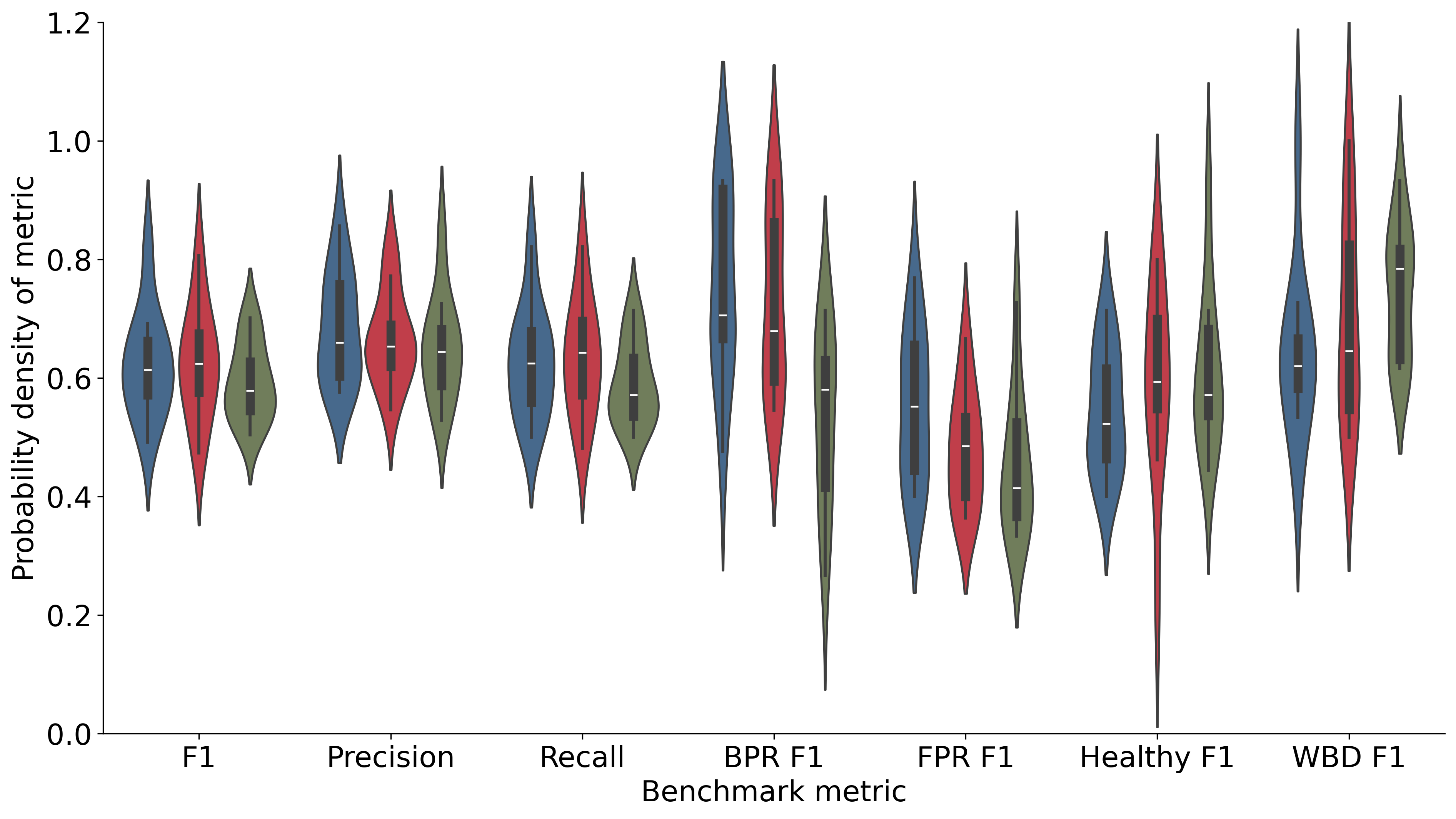}
    \caption{Validation dataset results}
    \label{fig:short-b}
  \end{subfigure}
  \caption{Violin and box plots of 10-fold cross-validation results for PhytNet and ResNet18 trained(a) and validated(b) on infrared or RGB images of cocoa disease. Shown here is the Gaussian density function, medium and interquartile ranges for mean F1, per class F1, precision and recall. PhytNet was trained only on IR data, while ResNet18 was trained on IR or RGB data. The datasets had four classes: Black pod rot, Frosty pod rot, Healthy and Witches' broom disease. n=70 images per class of early to mid-stage diseased or healthy cocoa with a 90\%:10\% train, validation split.}
        \label{fig:cross-validation results}
\end{figure*}

PhytNet and ResNet18 had almost perfectly consistent mean training and validation F1 and loss scores, with PhytNet providing the lowest mean training and validation loss values. However, as we will now discuss, these simple metrics are hiding naive behaviour on the part of ResNet18. \Cref{fig:Grad-CAM attention maps} shows that ResNet18 is focusing its attention slightly better than EfficientNet, while PhytNet is focusing its attention on disease symptoms exceptionally well. We also see in  both FPR images of \Cref{fig:Grad-CAM attention maps}, that all four models seem to show some signs of being "distracted" or "overwhelmed" by the bright sunlight in the images, despite all four models making the correct classification in the first FPR image. However, despite this apparent distraction, PhytNet still shows that it focuses its attention on the cocoa pod or disease lesion.  

\Cref{fig:cross-validation results} shows, with greater detail than \Cref{tab:CrossVal_results}, the results from the cross-validation analysis of PhytNet trained on IR images and ResNet18 trained on IR or RGB images. These two model architectures are compared here as they fit best to the data according to the summary statistics in \Cref{tab:CrossVal_results} and the activation maps in \Cref{fig:Grad-CAM attention maps}. ResNet18 was chosen to compare training on the IR and RGB datasets as it is the better known and tested architecture. 

\Cref{fig:cross-validation results} shows that the median values per metric are very similar between models, with the exception of PhytNet's slightly higher validation F1 for FPR. ResNet18 seems to have a higher median validation F1 for Healthy but with very long tails for this class and WBD, meaning that its performance on the validation set was highly variable for these classes. Additionally, we see that the distribution of ResNet18 F1 values across classes for the training set are much narrower than in the validation set, with an almost perfect Gaussian distribution for training BPR F1. This indicates some degree of overfitting. \emph{i.e.} ResNet18 is learning the training data very well but struggles to generalise consistently across the validation set.

Comparing the PhytNet training and validation results, the relatively Gaussian shapes of the distributions for F1, precision, and recall suggest that PhytNet is providing consistent and reliable results across the dataset. This consistency is further reinforced by the similarity between training and validation distributions. The wide tails in the PhytNet per class F1 scores are a concern as they indicate that its performance for specific classes is quite variable. However, considering the very small size of the dataset used here (per class train n = 63, val n=7), this variability in model performance is likely due to inconsistencies and noise in the data for specific classes, affecting the model's performance. The identical distributions between training and validation set for PhytNet suggests that the model is not memorising the training data but is instead learning genuine patterns that are applicable to unseen data.

Additionally, the ResNet18 mean training F1 for FPR was 13\% \textbf{lower} than the mean FPR validation F1. This curious behaviour, in conjunction with the failure of ResNet18 to consistently isolate features of interest in the Grad-CAM analysis and the inconsistent distributions of values between train and validation sets suggest that ResNet18 is overfitting, if to a lesser extent than the EfficientNet variants.

\Cref{fig:cross-validation results} also shows a slight improvement in FPR classification using the IR images over the RGB images. This, in conjunction with the 80\% FPR accuracy of the random forest, corroborates the importance of the dip in IR reflectance of FPR pods seen in \Cref{fig:spectroscopy-a}. Though it should be noted the BPR and Healthy accuracy was equally high with the random forest, despite no such obvious signal in the reflectance spectrum. We also see in \Cref{fig:cross-validation results} that the BPR F1 distribution was markedly lower when using the RGB dataset, while median WBD F1 was much greater.

\section{DISCUSSION}
\label{sec:sample3}
\subsection{Photosynthetic Activity as a Disease Indicator}
We observed no significant changes in photosynthetic yield or non-photochemical quenching between healthy leaves and those adjacent to diseased pods or leaves. This suggests that, if there exist systemic effects of these diseases on photosynthesis, they are not readily detectable, at least in the early stages. This result reinforces the need to focus on localised symptoms for disease detection in cocoa, which is consistent with what we know of the pathology of these diseases. \emph{i.e.} While \emph{Phytophthora spp.} can cause seedling blight and trunk cankers in adult trees, BPR and FPR symptoms tend to be isolated to the infected pods \cite{kraussMoniliophthoraRoreriFrosty2012, cabiPhytophthoraMegakaryaBlack2021a}. However, the sample size of this experiment is limiting and so, although the effect sizes seem to demonstrate a clear pattern, these results should be considered accordingly.

\subsection{Spectral Characteristics of Cocoa Pod Diseases}
The reduction in infrared light reflectance in FPR infected pods observed here could be an important finding. This is because FPR often remains asymptomatic during its biotrophic phase, until the later stages of disease progression \cite{baileyMoniliophthoraRoreriCausal2018b}, making early detection exceedingly difficult. This reduced IR reflectance may be indicative of altered metabolic activity within a pod resulting from FPR infection. This is in contrast to BPR, which primarily affects the external pod surface, leaving the internal metabolic activities relatively unaltered until the later stages of disease progression \cite{akrofiPhytophthoraMegakaryaReview2015}.

The spectroscopy data proved to be substantially informative for disease classification, especially in the UV and red-IR range. The feature importance scores presented here highlight that the most informative spectra lie in the UVA and red-infrared ranges, which are not typically captured in standard RGB images. This suggests that computer vision models relying solely on RGB images may face challenges in accurate disease classification and should incorporate additional spectral data or rely more on morphological characteristics. This result is somewhat corroborated by the observation that ResNet18 trained on IR images had slightly better median per class BPR and FPR F1 validation scores than when trained on RGB data.

While we randomised across factors such as disease stage and crop variety, a more controlled analysis should be undertaken to confirm these findings. A study that more carefully controlled for factors such as air temperature , internal pod temperature and crop variety differences would be highly informative.

An additional informative feature for disease detection via computer vision that should be considered in future studies is polarised light. Polarised light can indicate plant cuticle properties related to stress and disease \cite{nilssonRemoteSensingImage1995, gunasekaranComputerVisionTechnology1996}. However, while polarised light has been used in conjunction with CNNs to successfully map rice paddies \cite{bemIrrigatedRiceCrop2021}, little such work appears to have been done since the advent of CNNs for plant disease detection.       

\subsection{Evaluation of Convolutional Neural Networks}
The EfficientNet variants gave high F1 scores but also showed clear signs of overfitting. In contrast, PhytNet and ResNet18 demonstrated more consistent performance in F1 scores and Grad-CAM analysis. PhytNet performed by far the best in the Grad-CAM analysis, while also having the lowest loss values and least GFLOPS. 

The effects of bright sunlight may partly explain why the median FPR F1 was relatively low in all models here. As the causative agent of FPR, \emph{M. perniciosa}, is wind dispersed \cite{leachModellingMoniliophthoraRoreri2002} FPR tends to be found above two meters in the canopy. This means that bright light will be more common in these images than in images of the soil-borne BPR\cite{nobleSuppressionSoilbornePlant2005}. This may suggest a need for controlled imaging conditions. However, the fact that all four models correctly classified the first FPR image, despite this apparent distraction, suggests that the models might still leverage other features for classification when bright light is present. While PhytNet focuses on the diseased pods in addition to bight light, the other three models focus on irrelevant features such as the tree trunk, background or only bright light. However, despite this, PhytNet classifies the second FPR image as healthy, possibly because only healthy pod tissue is brightly illuminated. 

While PhytNet had the lowest mean validation F1 score, it is the only model here that showed no signs of overfitting, it performed with great consistency between train and validation sets, and it performed best for FPR images. This later point being of high potential economic value to cocoa farmers. As such, PhytNet would most likely perform best and most consistently in the field.   

\subsection{Optimisation of output node number}
During the optimisation sweep, PhytNet performed consistently better with seven or eight output nodes, despite having only four classes in the dataset. In challenging classification problems, having more output nodes than classes offers several potential advantages. Extra nodes may capture nuanced feature representations \cite{donahueAdversarialFeatureLearning2017}, act as "soft clusters" for variations within classes, or serve as a form of regularisation to improve generalisation though increased complexity of model parameters \cite{devriesImprovedRegularizationConvolutional2017}. The extra nodes could also provide a "catch-all" for unknown or less frequent features, thereby preventing forced guesses and acting to prevent representation collapse, where disparate inputs are mapped to the same point or a very narrow region in the feature space  \cite{goyalDROCCDeepRobust2020}. Additionally, a larger parameter space may smooth the optimisation landscape, a characteristic that is said to make it easier for algorithms to find a good solution \cite{walshawMultilevelLandscapesCombinatorial2002, kellScientificDiscoveryCombinatorial2012a}. These potential explanations for this curious model behaviour should be explored in future work.

\section{Conclusion}
PhytNet, while promising, has potential limitations that have yet to be tested. In future studies we will test if its performance is dataset-dependent and how well it  captures intricate patterns in more complex data. Although efficient, its reduced complexity might be a compromising factor, and it could be prone to underfitting in certain scenarios. We will test PhytNet's ability in transfer learning in a future study, though with a larger plant pathology dataset rather than ImageNet, which is irrelevant in this context. PhytNet emerged as a promising candidate model architecture in our study, particularly in its ability to focus its attention exceptionally well on relevant features like cocoa pods and disease lesions. Additionally, PhytNet is approximately 5 times faster than ResNet18 at inference time. The superior attention of PhytNet and apparent complete lack of overfitting offers unique advantages that could be leveraged for specific applications, such as localisation of disease symptoms on a tree. For example, in automated fungicide application systems, the ability to accurately pinpoint the location of the disease symptom or pathogen could lead to more efficient and targeted application of fungicides, thereby reducing waste and pollution.
{
    \small
    \bibliographystyle{ieeenat_fullname}
    \bibliography{main}
}


\end{document}